\documentclass[11pt,letterpaper]{article}
\usepackage{times}
\usepackage{url}
\usepackage{latexsym}
\usepackage{subcaption}
\usepackage{graphicx}
\usepackage{placeins}
\usepackage{hyperref}
\usepackage{minibox}
\usepackage{emnlp2016}

\DeclareCaptionFont{tiny}{\tiny}

\usepackage{amsmath,amsthm,amssymb,amsfonts,mathtools}

\usepackage{tabularx}
\usepackage{multirow}
\usepackage{wrapfig}

\usepackage[justification=justified,singlelinecheck=false]{caption}

\usepackage{algorithm}
\usepackage{algpseudocode}

\usepackage{tikz-dependency}

\newcommand{\textoverline}[1]{$\overline{\mbox{#1}}$}

\newcommand{\APT}{%
	\textsc{Apt}}

\emnlpfinalcopy

\def\emnlppaperid{817}

\title{Improving Sparse Word Representations with Distributional Inference for Semantic Composition}

\author{
  Thomas Kober, Julie Weeds, Jeremy Reffin \and David Weir\\
 TAG laboratory, Department of Informatics, University of Sussex\\
 Brighton, BN1 9QH, UK\\
  {\tt \{t.kober, j.e.weeds, j.p.reffin, d.j.weir\}@sussex.ac.uk}
}

\date{}
\begin{document}
\maketitle

\begin{abstract}
Distributional models are derived from co-occurrences in a corpus, where only a small proportion of all possible plausible co-occurrences will be observed. This results in a very sparse vector space, requiring a mechanism for inferring missing knowledge. Most methods face this challenge in ways that render the resulting word representations uninterpretable, with the consequence that semantic composition becomes hard to model. In this paper we explore an alternative which involves explicitly inferring unobserved co-occurrences using the distributional neighbourhood. We show that distributional inference improves sparse word representations on several word similarity benchmarks and demonstrate that our model is competitive with the state-of-the-art for adjective-noun, noun-noun and verb-object compositions while being fully interpretable.
\end{abstract}

\section{Introduction}
The aim of distributional semantics is to derive meaning representations based on observing co-occurrences of words in large text corpora. However not all plausible co-occurrences will be observed in any given corpus, resulting in word representations that only capture a fragment of the meaning of a word. For example the verbs ``walking" and ``strolling" may occur in many different and possibly disjoint contexts, although both verbs would be equally plausible in numerous cases. This subsequently results in incomplete representations for both lexemes. In addition, models based on counting co-occurrences face the general problem of sparsity in a very high-dimensional vector space. The most common approaches to these challenges have involved the use of various techniques for dimensionality reduction~\cite{Bullinaria_2012,Lapesa_2014} or the use of low-dimensional and dense neural word embeddings~\cite{Mikolov_2013a,Pennington_2014}. The common problem in both of these approaches is that composition becomes a black-box process due to the lack of interpretability of the representations. Count-based models are therefore a very attractive line of work with regards to a number of important long-term research challenges, most notably the development of an adequate model of distributional compositional semantics. In this paper we propose the use of distributional inference (DI) to inject unobserved but plausible distributional semantic knowledge into the vector space by leveraging the intrinsic structure of the distributional neighbourhood. This results in richer word representations and furthermore mitigates the sparsity effect common in high-dimensional vector spaces, while remaining fully interpretable.\\
Our contributions are as follows: we show that typed and untyped sparse word representations, enriched by distributional inference, lead to performance improvements on several word similarity benchmarks, and that a higher-order dependency-typed vector space model, based on ``Anchored Packed Dependency Trees ({\APT}s)" \cite{Weir_2016}, is competitive with the state-of-the-art for adjective-noun, noun-noun and verb-object compositions. Using our method, we are able to bridge the gap in performance between high dimensional interpretable models and low dimensional non-interpretable models and offer evidence to support a possible explanation of why high-dimensional models usually perform worse, together with a simple, practical method for over-coming this problem. We furthermore demonstrate that \emph{intersective} approaches to composition benefit more from distributional inference than composition by \emph{union} and highlight the ability of composition by \emph{intersection} to disambiguate the meaning of a phrase in a local context.\\
The remainder of this paper is structured as follows: we discuss related work in section~\ref{related_work}, followed by an introduction of the \APT~framework for semantic composition in section~\ref{background}. We describe distributional inference in section~\ref{approach} and present our experimental work, together with our results in section~\ref{sec:experiments}. We conclude this paper and outline future work in section~\ref{conclusion}.
\label{introduction}

\section{Related Work}
Our method follows the distributional smoothing approach of Dagan et al.~\shortcite{Dagan_1994} and Dagan et al.~\shortcite{Dagan_1997}. In these works the authors are concerned with smoothing the probability estimate for unseen words in bigrams. This is achieved by measuring which unobserved bigrams are more likely than others on the basis of the Kullback-Leibler divergence between bigram distributions. This has led to significantly improved performance on a language modelling for speech recognition task, as well as for word-sense disambiguation in machine translation~\cite{Dagan_1994,Dagan_1997}. More recently Pad\'{o} et al.~\shortcite{Pado_2013} used a distributional approach for smoothing derivationally related words, such as \emph{oldish -- old}, as a back-off strategy in case of data sparsity. However, none of these approaches have used distributional inference as a general technique for directly enriching sparse distributional vector representations, or have explored its behaviour for semantic composition.\\
Compositional models of distributional semantics have become an increasingly popular topic in the research community. Starting from simple pointwise additive and multiplicative approaches to composition, such as Mitchell and Lapata~\shortcite{Mitchell_2008,Mitchell_2010}, and Blacoe and Lapata~\shortcite{Blacoe_2012}, to tensor based models, such as Baroni and Zamparelli~\shortcite{Baroni_2010a}, Coecke et al.~\shortcite{Coecke_2010}, Grefenstette et al.~\shortcite{Grefenstette_2013} and Paperno et al.~\shortcite{Paperno_2014}, and neural network based approaches, such as Socher et al.~\shortcite{Socher_2012}, Le and Zuidema~\shortcite{Le_2015}, Mou et al.~\shortcite{Mou_2015} and Tai et al.~\shortcite{Tai_2015}. Zanzotto et al.~\shortcite{Zanzotto_2015} provide a decompositional analysis of how similarity is affected by distributional composition, and link compositional models to convolution kernels. Most closely related to our approach of composition are the works of Thater et al.~\shortcite{Thater_2010}, Thater et al.~\shortcite{Thater_2011} and Weeds et al.~\shortcite{Weeds_2014}, which aim to provide a general model of compositionality in a typed distributional vector space. In this paper we adopt the approach to distributional composition introduced by Weir et al.~\shortcite{Weir_2016}, whose \APT~framework is based on a higher-order dependency-typed vector space, however they do not address the issue of sparsity in their work.
\label{related_work}

\section{Background}
\begin{figure}[!htb]
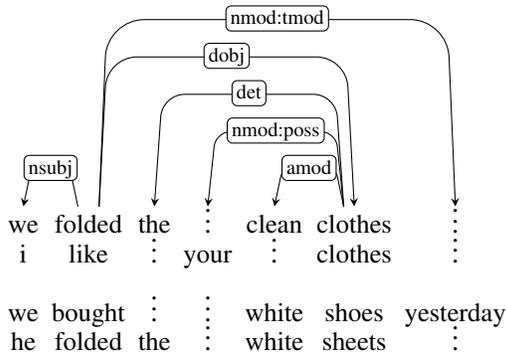

\setlength{\belowcaptionskip}{-10pt}
\setlength{\abovecaptionskip}{-5pt}
\centering
\begin{dependency}
\begin{deptext}
  \small{we}\&
    \small{folded}\& 
      \small{the}\&
        {$\vdots$}\&
          \small{clean}\&
            \small{clothes}\&
              {$\vdots$}\\
   \small{i}\&
    \small{like}\& 
      {$\vdots$}\&
        \small{your}\&
          {$\vdots$}\&
            \small{clothes}\&
              {$\vdots$}\\\\
  
  \small{we}\&
    \small{bought}\&
      {$\vdots$}\&
        {$\vdots$}\&
          \small{white}\&
            \small{shoes}\&
              \small{yesterday}\\
   \small{he}\&
    \small{folded}\& 
      \small{the}\&
        {$\vdots$}\&
          \small{white}\&
            \small{sheets}\&
              {$\vdots$}\\
\end{deptext}
\depedge{2}{1}{nsubj}
\depedge{6}{5}{amod}
\depedge{6}{3}{det}
\depedge{2}{6}{dobj}
\depedge{2}{7}{nmod:tmod}
\depedge{6}{4}{nmod:poss}
\end{dependency}
\caption{Aligned Packed Dependency Tree representation of the example sentences. }
\label{fig:apt}
\end{figure}

Distributional vector space models can broadly be categorised into untyped proximity based models and typed models \cite{Baroni_2010b}. Examples of the former include Deerwester et al.~\shortcite{Deerwester_1990}; Lund and Burgess~\shortcite{Lund_1996}; Curran~\shortcite{Curran_2004}; Sahlgren~\shortcite{Sahlgren_2006}; Bullinaria and Levy~\shortcite{Bullinaria_2007} and Turney and Pantel~\shortcite{Turney_2010}. These models count the number of times every word in a large corpus co-occurs with other words within a specified spatial context window, without leveraging the structural information of the text. Typed models on the other hand, take the grammatical relation between two words for a co-occurrence event into account. Early proponents of that approach are Grefenstette~\shortcite{Grefenstette_1994} and Lin~\shortcite{Lin_1998}. More recent work by Pad\'{o} and Lapata~\shortcite{Pado_2007}, Erk and Pad\'{o}~\shortcite{Erk_2008} and Weir et al.~\shortcite{Weir_2016} uses dependency paths to build a structured vector space model. In both kinds of models, the raw counts are usually transformed by Positive Pointwise Mutual Information (PPMI) or a variant of it \cite{Church_1990,Niwa_1994,Scheible_2013,Levy_2014c}.\\
In the following we will give an explanation of the theory of composition with {\APT}s as introduced by Weir et al.~\shortcite{Weir_2016}, which we adopt in this paper. In addition to direct relations between two words, the \APT~model also considers \emph{inverse} and \emph{higher order} relations. Inverse relations are denoted with a horizontal bar above the dependency relation, such as \texttt{\textoverline{amod}} for an inverse adjectival modifier. Higher order dependencies are separated by a colon as in the second order distributional feature \texttt{\textoverline{dobj}:nsubj}. The example below illustrates how raw text is processed to retrieve elementary representations in our \APT~model. As an example we consider a lowercased corpus consisting of the sentences:\\
\makebox[\columnwidth]{\minibox{\emph{we folded the clean clothes}\\\emph{i like your clothes}\\\emph{we bought white shoes yesterday}\\\emph{he folded the white sheets}}} \\\\
We dependency parse the raw sentences and, following Weir et al.~\shortcite{Weir_2016}, align and aggregate the resulting parse trees according to their dependency type as shown in Figure~\ref{fig:apt}. For example the lexeme \emph{clothes} has the distributional features \texttt{amod:}\emph{dry} and \texttt{\textoverline{dobj}:nsubj:}\emph{we} among others. Over a large corpus, this results in a very high-dimensional and sparse vector space, which due to its typed nature is much sparser than for untyped models.
\subsubsection*{Composition with {\APT}s}
\noindent Composition is linguistically motivated by the principle of compositionality, which states that the meaning of a complex expression is fully determined by its structure and the meanings of its constituents \cite{Frege_1884}. Many simple approaches to semantic composition neglect the structure and lose information in the composition process. For example, the phrases \emph{house boat} and \emph{boat house} have the exact same representation when composition is done via a pointwise arithmetic operation. Despite performing well in a number of studies, this commutativity is not desirable for a fine grained understanding of the semantics of natural language.
When performing composition with {\APT}s, we adopt the method introduced by Weir et al.~\shortcite{Weir_2016} which views distributional composition as a process of contextualisation. 
\begin{table*}[!htb]
\centering
\small
\resizebox{\textwidth}{!}{
\begin{tabular}{ l | l | c || l | c }
\multicolumn{3}{c||}{\textbf{\emph{white}}} & \multicolumn{2}{c}{\textbf{\emph{clothes}}} \\
\textbf{Distributional Features} & \textbf{Offset Features (by \texttt{amod})} & \textbf{Co-occurrence Count} & \textbf{Distributional Features} & \textbf{Co-occurrence Count} \\ \hline
\texttt{\textoverline{amod}:}\emph{shoes} & \texttt{:}\emph{shoes} & $1$ & \texttt{amod:}\emph{clean} & $1$ \\
\texttt{\textoverline{amod}:\textoverline{dobj}:}\emph{bought} & \texttt{\textoverline{dobj}:}\emph{bought} & $1$ & \texttt{\textoverline{dobj}:}\emph{like} & $1$ \\
\texttt{\textoverline{amod}:\textoverline{dobj}:}\emph{folded} & \texttt{\textoverline{dobj}:}\emph{folded} & $1$ & \texttt{\textoverline{dobj}:}\emph{folded} & $1$ \\
\texttt{\textoverline{amod}:\textoverline{dobj}:nsubj:}\emph{we} & \texttt{\textoverline{dobj}:nsubj:}\emph{we} & $1$ & \texttt{\textoverline{dobj}:nsubj:}\emph{we} & $1$ \\
\end{tabular}}
\captionsetup{font=tiny}
\caption{Example feature spaces for the lexemes \emph{white} and \emph{clothes} extracted from the dependency tree of Figure~\ref{fig:apt}. Not all features are displayed for space reasons. Offsetting \texttt{\textoverline{amod}:}\emph{shoes} by \texttt{amod} results in an empty dependency path, leaving just the word co-occurrence \texttt{:}\emph{shoes} as feature.}
\label{feature_space_example}
\end{table*}
For composing the adjective \emph{white} with the noun \emph{clothes} via the dependency relation \texttt{amod} we need to consider how the adjective interacts with the noun in the vector space. The distributional features of white describe things that are white via their first order relations such as \texttt{\textoverline{amod}}, and things that can be done to white things, such as \emph{bought} via \texttt{\textoverline{amod}:\textoverline{dobj}} in the example above. \\
Table~\ref{feature_space_example} shows a number of features extracted from the aligned dependency trees in Figure~\ref{fig:apt} and highlights that adjectives and nouns do not share many features if only first order dependencies would be considered. However through the inclusion of inverse and higher order dependency paths we can observe that the second order features of the adjective align with the first order features of the noun. 
\begin{table*}[!htb]
\centering
\small
\begin{tabular}{ l | c || l | c }
\multicolumn{2}{c||}{\textbf{Composition by \emph{union}}} & \multicolumn{2}{c}{\textbf{Composition by \emph{intersection}}} \\
\textbf{Distributional Features} & \textbf{Co-occurrence Count} & \textbf{Distributional Features} & \textbf{Co-occurrence Count} \\ \hline
\texttt{:}\emph{shoes} & $1$ &  & \\
\texttt{amod:}\emph{clean} & $1$ &  &  \\
\texttt{\textoverline{dobj}:}\emph{bought} & $1$ & & \\
\texttt{\textoverline{dobj}:}\emph{folded} & $2$ & \texttt{\textoverline{dobj}:}\emph{folded} & $2$ \\
\texttt{\textoverline{dobj}:}\emph{like} & $1$ && \\
\texttt{\textoverline{dobj}:nsubj:}\emph{we} & $2$ & \texttt{\textoverline{dobj}:nsubj:}\emph{we} & $2$ \\
\end{tabular}
\captionsetup{font=tiny}
\caption{Comparison of composition by \emph{union} and composition by \emph{intersection}. Not all features are displayed for space reasons.}
\label{composition_example}
\end{table*}
For composition, the adjective \emph{white} needs to be offset by its \emph{inverse} relation to \emph{clothes}\footnote{The inverse of \texttt{\textoverline{amod}} is just \texttt{amod}.} making it distributionally similar to a noun that has been modified by \emph{white}. Offsetting can be seen as shifting the current viewpoint in the \APT~data structure and is necessary for aligning the feature spaces for composition~\cite{Weir_2016}. We are then in a position to compose the offset representation of \emph{white} with the vector for \emph{clothes} by the \emph{union} or the \emph{intersection} of their features. \\
Table~\ref{composition_example} shows the resulting feature spaces of the composed vectors. It is worth noting that any arithmetic operation can be used to combine the counts of the aligned features, however for this paper we use pointwise addition for both composition functions. 
One of the advantages of this approach to composition is that the inherent interpretability of count-based models naturally expands beyond the word level, allowing us to study the distributional semantics of phrases in the same space as words. Due to offsetting one of the constituents, the composition operation is not commutative and hence avoids identical representations for \emph{house boat} and \emph{boat house}. However, the typed nature of our vector space results in extreme sparsity, for example while the untyped VSM has $130$k dimensions, our \APT~model can have more than $3$m dimensions. We therefore need to enrich the elementary vector representations with the distributional information of their nearest neighbours to ease the sparsity effect and infer missing information. Due to the syntactic nature of our composition operation it is not straightforward to apply common dimensionality reduction techniques such as SVD, as the type information needs to be preserved.
\label{background}

\section{Distributional Inference}
Following Dagan et al.~\shortcite{Dagan_1994} and Dagan et al.~\shortcite{Dagan_1997}, we propose a simple unsupervised algorithm for enriching sparse vector representations with their nearest neighbours. We show that our distributional inference algorithm improves performance for untyped and typed models on several word similarity benchmarks, as well as being competitive with the state-of-the-art on semantic composition. 
As shown in algorithm~\ref{alg:distributional_inference} below, we iterate over all word vectors $w$ in a given distributional model $M$, and add the vector representations of the nearest neighbours $n$, determined by cosine similarity, to the representation of the enriched word vector $w^{\prime}$. The parameter $\alpha$ in line $4$ scales the contribution of the original word vector to the resulting enriched representation. In this work we always chose $\alpha$ to be identical to the number of neighbours used for distributional inference. For example, if we used $10$ neighbours for DI, we would set $\alpha=10$, which we found sufficient to prevent the neighbours from dominating the vector representation. In our experiments we kept the input distributional model fixed, however it is equally possible to update the given model in an online fashion, adding some amount of stochasticity to the enriched word vector representations. There is a number of possibilities for the neighbour retrieval function $neighbours()$ and we explore several options in this paper. The algorithm furthermore is agnostic to the input distributional model, for example it is possible to use completely different vector space models for querying neighbours and enrichment. 
\begin{algorithm}
	\caption{Distributional Inference}
	\label{alg:distributional_inference}
	\begin{algorithmic}[1]
		\Procedure{dist\_inference}{$M$}
			\State \textbf{init} $M^{\prime}$
			\ForAll{$w$ in $M$}
				\State $w^{\prime} \gets w \times \alpha$
				\ForAll{$n$ in $neighbours(M, w)$}
					\State $w^{\prime} \gets w^{\prime} + n$
					\State \textbf{add} $w^{\prime}$ \textbf{to} $M^{\prime}$
				\EndFor
			\EndFor
			\State \textbf{return} $M^{\prime}$
		\EndProcedure
	\end{algorithmic}
\end{algorithm}
\vspace*{-0.4cm}
\subsubsection*{Static Top $n$ Neighbour Retrieval}
\noindent The perhaps simplest way is to choose the top $n$ most similar neighbours for each word in the vector space and enrich the respective vector representations with them.

\subsubsection*{Density based Neighbour Retrieval}
\noindent This approach has its roots in kernel density estimation~\cite{Parzen_1962}, however instead of defining a static global parzen window, we set the window size for every word individually, depending on the distance to its nearest neighbour, plus a threshold. For example if the cosine distance between the target vector and its top neighbour is $0.5$, we use a window size of $0.5 + \epsilon$ for that word. In our experiments we typically define $\epsilon$ to be proportional to the distance of the nearest neighbour (e.g. $\epsilon = 0.5 \times 0.1$).

\subsubsection*{WordNet based Neighbour Retrieval}
\noindent Instead of leveraging the intrinsic structure of our distributional vector space, we retrieve neighbours by querying WordNet \cite{Fellbaum_1998}, and treat synsets with agreeing PoS tags as the nearest neighbours of any target vector. This restricts the retrieved neighbours to synonyms only.
\label{approach}

\section{Experiments}
\label{sec:experiments}
Our model is based on a cleaned October 2013 Wikipedia dump, which excludes all pages with fewer than $20$ page views, resulting in a corpus of approximately $0.6$ billion tokens \cite{Wilson_2015}. The corpus is lowercased, tokenised, lemmatised, PoS tagged and dependency parsed with the Stanford NLP tools, using universal dependencies \cite{Manning_2014,deMarneffe_2014}. We then build our \APT~model with first, second and third order relations. We remove distributional features with a count of less than $10$, and vectors containing fewer than $50$ non-zero entries. The raw counts are subsequently transformed to PPMI weights. The untyped vector space model is built from the same lowercased, tokenised and lemmatised Wikipedia corpus. We discard terms with a frequency of less than $50$ and apply PPMI to the raw co-occurrence counts.

\subsubsection*{Shifted PPMI}
\noindent We explore a range of different values for shifting the PPMI scores as these have a significant impact on the performance of the \APT~model. The effect of shifting PPMI scores for untyped vector space models has already been explored in Levy and Goldberg~\shortcite{Levy_2014c}, and Levy et al.~\shortcite{Levy_2015}, thus we only present results for the \APT~model.  As shown in equation~\ref{pmi}, PMI is defined as the log of the ratio of the joint probability of observing a word $w$ and a context $c$ together, and the product of the respective marginals of observing them separately. In our \APT~model, a context $c$ is defined as a dependency relation together with a word. 
\setlength{\belowdisplayskip}{0pt}
\setlength{\abovedisplayskip}{0pt}
\setlength{\belowdisplayshortskip}{0pt}
\setlength{\abovedisplayshortskip}{0pt}
\begin{equation}
\small
\label{pmi}
\begin{split}
PMI(w, c) &= \log{\frac{P(w, c)}{P(w)P(c)}}\\
SPPMI(w, c) &= \max(PMI(w, c) - \log k, 0)
\end{split}
\normalsize
\end{equation}
As PMI is negatively unbounded, PPMI is used to ensure that all values are greater than or equal to $0$.  Shifted PPMI (SPPMI) subtracts a constant from any PMI score before applying the PPMI threshold. We experiment with values of $1$, $5$, $10$, $40$ and $100$ for the shift parameter $k$. 

\subsection{Word Similarity Experiments}

\begin{figure}[!htb]
\setlength{\belowcaptionskip}{-5pt}
\setlength{\abovecaptionskip}{-10pt}
\includegraphics[width=\columnwidth]{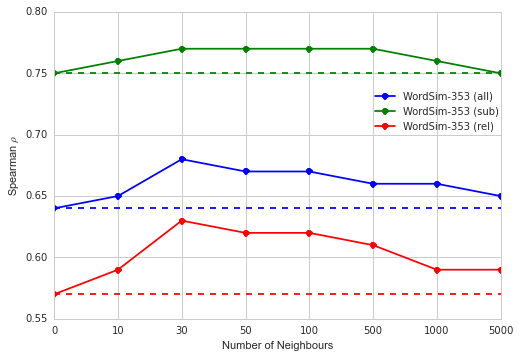}
\captionsetup{font=tiny}
\caption{Effect of the number of neighbours on WordSim-353.}
\label{effect_of_neighbours}
\end{figure}

We first evaluate our models on $3$ word similarity benchmarks, MEN~\cite{Bruni_2014}, which is testing for \emph{relatedness} (e.g. meronymy or holonymy) between terms, SimLex-999~\cite{Hill_2015}, which is testing for \emph{substitutability} (e.g. synonymy, antonymy, hyponymy and hypernymy), and WordSim-353~\cite{Finkelstein_2001}, where we use the version of Agirre et al.~\shortcite{Agirre_2009}, who split the dataset into a \emph{relatedness} and a \emph{substitutability} subset. Baroni and Lenci~\shortcite{Baroni_2011} have shown that untyped models are typically better at capturing \emph{relatedness}, whereas typed models are better at encoding \emph{substitutability}. Performance is measured by computing Spearman's $\rho$ between the cosine similarities of the vector representations and the corresponding aggregated human similarity judgements. For these experiments we keep the number of neighbours that a word vector can consume fixed at  $30$. This value is based on preliminary experiments on WordSim-353 (see Figure~\ref{effect_of_neighbours}) using the static top $n$ neighbour retrieval function and a PPMI shift of $k=40$. Figure~\ref{effect_of_neighbours} shows that distributional inference improves performance for any number of neighbours over a model without DI (marked as horizontal dashed lines for each WordSim-353 subset) and peaks at a value of $30$. Performance slightly degrades with more neighbours. For the untyped VSM we use a symmetric window of $5$ on either side of the target word.
\begin{table*}[!htb]
\centering
\small
\begin{tabular}{ l | c | c | c | c | c | c | c | c }
\multirow{2}{*}{\bf{{\APT}s}} & \multicolumn{2}{ c |}{\bf{MEN}} & \multicolumn{2}{ c |}{\bf{SimLex-999}} & \multicolumn{2}{ c |}{\bf{WordSim-353 (rel)}} & \multicolumn{2}{ c }{\bf{WordSim-353 (sub)}}\\ 
& \emph{without DI} & \emph{with DI} & \emph{without DI} & \emph{with DI} & \emph{without DI} & \emph{with DI} & \emph{without DI} & \emph{with DI}\\ \hline
\emph{$k=1$} & $0.54$ & $0.52$ & $0.31$ & $0.30$ & $0.34$ & $0.27$ & $0.62$ & $0.60$ \\ 
\emph{$k=5$} & $0.64$ & $0.65$ & $0.35$ & $\bf{0.36}$ & $0.56$ & $0.51$ & $0.74$ & $0.73$ \\
\emph{$k=10$} & $0.63$ & $0.66$ & $0.35$ & $\bf{0.36}$ & $0.56$ & $0.55$ & $0.75$ & $0.74$ \\ 
\emph{$k=40$} & $0.63$ & $\bf{0.68}$ & $0.30$ & $0.32$ & $0.55$ & $\bf{0.61}$ & $0.75$ & $\bf{0.76}$ \\ 
\emph{$k=100$} & $0.61$ & $0.67$ & $0.26$ & $0.29$ & $0.47$  & $0.60$  & $0.71$  & $0.72$\\ \hline
\end{tabular}
\captionsetup{singlelinecheck=false,font=tiny,labelsep=newline}
\caption{Effect of the magnitude of the shift parameter $k$ in SPPMI on the word similarity tasks. Boldface means best performance per dateset.}
\label{wordsim_results_shift}
\end{table*}

Table~\ref{wordsim_results_shift} highlights the effect of the SPPMI shift parameter $k$, while keeping the number of neighbours fixed at $30$ and using the static top $n$ neighbour retrieval function. For the \APT~model, a value of $k=40$ performs best (except for SimLex-999, where smaller shifts give better results), with a performance drop-off for larger shifts. In our experiments we find that a shift of $k=1$ results in top performance for the untyped vector space model. It appears that shifting the PPMI scores in the \APT~model has the effect of cleaning the vectors from noisy PPMI artefacts, which reinforces the predominant sense, while other senses get suppressed. Subsequently, this results in a cleaner neighbourhood around the word vector, dominated by a single sense. This explains why distributional inference slightly degrades performance for smaller values of $k$.
\begin{table*}[!htb]
\centering
\small
\begin{tabular}{ l | c | c | c | c }
\bf{{\APT}s ($k=40$)} & \bf{No Distributional Inference} & \bf{Density Window} & \bf{Static Top $n$} & \bf{WordNet} \\ 
\emph{MEN} & $0.63$ & $0.67$ & $\bf{0.68}$ & $0.63$ \\ 
\emph{SimLex-999} & $0.30$ & $0.32$ & $0.32$ & $\bf{0.38}$ \\ 
\emph{WordSim-353 (rel)} & $0.55$ & $\bf{0.62}$ & $0.61$ & $0.56$ \\ 
\emph{WordSim-353 (sub)} & $0.75$ & $\bf{0.78}$ & $0.76$ & $0.77$ \\ \hline
\bf{Untyped VSM ($k=1$)} & \bf{No Distributional Inference} & \bf{Density Window} & \bf{Static Top $n$} & \bf{WordNet} \\ 
\emph{MEN*} & $\bf{0.71}$ & $\bf{0.71}$ & $\bf{0.71}$ & $\bf{0.71}$ \\ 
\emph{SimLex-999} & $0.30$ & $0.29$ & $0.30$ & $\bf{0.36}$ \\ 
\emph{WordSim-353 (rel)} & $0.60$ & $\bf{0.64}$ & $\bf{0.64}$ & $0.52$ \\ 
\emph{WordSim-353 (sub)} & $0.70$ & $\bf{0.73}$ & $0.72$ & $0.67$ \\ \hline
\end{tabular}
\captionsetup{font=tiny}
\caption{Neighbour retrieval function comparison. Boldface means best performance on a dataset \emph{per} VSM type. *) With 3 significant figures, the density window approach ($0.713$) is slightly better than the baseline without DI ($0.708$), static top $n$ ($0.710$) and WordNet ($0.710$).}
\label{wordsim_results_neighbour_fn}
\end{table*}

Table~\ref{wordsim_results_neighbour_fn} shows that distributional inference successfully infers missing information for both model types, resulting in improved performance over models without the use of DI on all datasets. The improvements are typically larger for the \APT~model, suggesting that it is missing more distributional knowledge in its elementary representations than untyped models. The density window and static top $n$ neighbour retrieval functions perform very similar, however the static approach is more consistent and never underperforms the baseline for either model type on any dataset. The WordNet based neighbour retrieval function performs particularly well on SimLex-999. This can be explained by the fact that antonyms, which frequently happen to be among the nearest neighbours in distributional vector spaces, are regarded as dissimilar in SimLex-999, whereas the WordNet neighbour retrieval function only returns synonyms. The results furthermore confirm the effect that untyped models perform better on datasets modelling \emph{relatedness}, whereas typed models work better for \emph{substitutability} tasks~\cite{Baroni_2011}. 

\subsection{Composition Experiments}

Our approach to semantic composition as described in section~\ref{background} requires the dimensions of our vector space models to be meaningful and interpretable. However, the problem of missing information is amplified in compositional settings as many compatible dimensions between words are not observed in the source corpus. It is therefore crucial that distributional inference is able to inject some of the missing information in order to improve the composition process. For the experiments involving semantic composition, we enrich the elementary representations of the phrase constituents before composition.\\
We first conduct a qualitative analysis for our \APT~model and observe the effect of distributional inference on the nearest neighbours of composed adjective-noun, noun-noun and verb-object compounds. In these experiments, we show how distributional inference changes the neighbourhood in which composed phrases are embedded, and highlight the difference between composition by \emph{union} and composition by \emph{intersection}. For this experiment we use the static top $n$ neighbour retrieval function with $30$ neighbours and $k=40$. \\
Table \ref{composition_results_qualitative} shows a small number of example phrases together with their top $3$ nearest neighbours, computed from the union of all words in the Wikipedia corpus and all phrase pairs in the Mitchell and Lapata~\shortcite{Mitchell_2010} dataset. As can be seen, nearest neighbours of phrases can be either single words or other composed phrases. Words or phrases marked with ``\textbf{*}" in Table~\ref{composition_results_qualitative} mean that DI introduced, or failed to downrank, a spurious neighbour, while boldface means that performing distributional inference resulted in a neighbourhood more coherent with the query phrase than without DI. \\
Table~\ref{composition_results_qualitative} shows that composition by \emph{union} is unable to downrank unrelated neighbours introduced by distributional inference. For example \emph{large quantity} is incorrectly introduced as a top ranked neighbour for the phrase \emph{small house}, due to the proximity of \emph{small} and \emph{large} in the vector space. The phrases \emph{market leader} and \emph{television programme} are two examples of incoherent neighbours, which the composition function was unable to downrank and where DI could not improve the neighbourhood. Composition by \emph{intersection} on the other hand vastly benefits from distributional inference. Due to the increased sparsity induced by the composition process, a neighbourhood without DI produces numerous spurious neighbours as in the case of the verb \emph{have} as a neighbour for \emph{win battle}. Distributional inference introduces qualitatively better neighbours for almost all phrases. For example, \emph{government leader} and \emph{opposition member} are introduced as top ranked neighbours for the phrase \emph{party leader}, and \emph{stress importance} and \emph{underline} are introduced as new top neighbours for the phrase \emph{emphasise need}. These results show that composition by \emph{union} does not have the ability to disambiguate the meaning of a word in a given phrasal context, whereas composition by \emph{intersection} has that ability but requires distributional inference to unleash its full potential.\\
\begin{table*}[!htb]
\centering
\small
\resizebox{\textwidth}{!}{
\begin{tabular}{ l | l | l | l | l | l }
\textbf{Phrase} & \textbf{Comp.} & \textbf{Union} & \textbf{Union (with DI)} & \textbf{Intersection} & \textbf{Intersection (with DI)}  \\ \hline
national & AN & government, regime, & government, regime\textbf{*}, & federal assembly, & federal assembly, government, \\ government && ministry & european state\textbf{*} & government, monopoly & local office\\ \hline
small & AN & house, public building, & house, public building, & apartment, cottage, & cottage, apartment, cabin\\ house && building & large quantity\textbf{*} & cabin &\\ \hline
party & NN & leader, market leader, & leader, government leader, & party official, NDP, & \textbf{government leader}, party official,\\ leader && government leader & market leader\textbf{*} & leader & \textbf{opposition member}\\ \hline
training & NN & programme, action programme, & programme, action programme\textbf{*}, & training college, trainee, & training college, \\ programme && television programme & television programme\textbf{*} & education course & education course, \textbf{seminar}\\ \hline
win battle & VO & win, win match, ties & win, win match, \textbf{fight war} & win match, win, have & \textbf{fight war}, \textbf{fight}, win match\\ \hline
emphasise & VO & emphasise, underline, & emphasise, underline, & emphasise, prioritize, & emphasise, \textbf{stress importance}, \\ need && underscore & underscore & negate & \textbf{underline}\\\hline
\end{tabular}}
\captionsetup{font=tiny}
\caption{Nearest neighbours AN, NN and VO pairs in the Mitchell and Lapata~\shortcite{Mitchell_2010} dataset, with and without distributional inference. Words and phrases marked with \textbf{*} denote spurious neighbours, boldfaced words and phrases mark improved neighbours.}
\label{composition_results_qualitative}
\end{table*}
For a quantitative analysis of distributional inference for semantic composition, we evaluate our model on the composition dataset of Mitchell and Lapata~\shortcite{Mitchell_2010}, consisting of $108$ adjective-noun, $108$ noun-noun, and $108$ verb-object pairs. The task is to compare the model's similarity estimates with the human judgements by computing Spearman's $\rho$. For comparing the performance of the different neighbour retrieval functions, we choose the same parameter settings as in the word similarity experiments ($k=40$ and using $30$ neighbours for DI).\\
\indent Table~\ref{composition_results_neighbour_fn} shows that the static top $n$ and density window neighbour retrieval functions perform very similar again. The density window retrieval function outperforms static top $n$ for composition by \emph{intersection} and \emph{vice versa} for composition by \emph{union}. The WordNet approach is competitive for composition by \emph{union}, but underperfoms the other approaches for composition by \emph{intersection} significantly. For further experiments we use the static top $n$ approach as it is computationally cheap and easy to interpret due to the fixed number of neighbours. Table~\ref{composition_results_neighbour_fn} also shows that while composition by \emph{intersection} is significantly improved by distributional inference, composition by \emph{union} does not appear to benefit from it.
\begin{table*}[!htb]
\centering
\small
\resizebox{\textwidth}{!}{
\begin{tabular}{ l | c | c | c | c | c | c | c | c }
\multirow{2}{*}{\bf{{\APT}s}} & \multicolumn{2}{c|}{\bf{No Distributional Inference}} & \multicolumn{2}{c|}{\bf{Density Window}} & \multicolumn{2}{c|}{\bf{Static Top $n$}} & \multicolumn{2}{c}{\bf{WordNet}} \\ 
& \emph{intersection} & \emph{union} & \emph{intersection} & \emph{union} & \emph{intersection} & \emph{union} & \emph{intersection} & \emph{union} \\ \hline
\emph{Adjective-Noun} & $0.10$ & \underline{$0.41$} & $0.31$ & $0.39$ & $0.25$ & $0.40$ & $0.12$ & \underline{$0.41$} \\
\emph{Noun-Noun} & $0.18$ & $0.42$ & $0.34$ & $0.38$ & $0.37$ & \underline{$0.45$} & $0.24$ & $0.36$ \\
\emph{Verb-Object} & $0.17$ & $\underline{0.36}$ & \underline{$0.36$} & \underline{$0.36$} & $0.34$ & $0.35$ & $0.25$ & \underline{$0.36$} \\ \hline
\bf{\emph{Average}} & $0.15$ & $\bf{0.40}$ & $0.34$ & $0.38$ & $0.32$ & $\bf{0.40}$ & $0.20$ & $0.38$ \\ \hline
\end{tabular}}
\captionsetup{font=tiny}
\caption{Neighbour retrieval function. Underlined means best performance per phrase type, boldface means best average performance overall.}
\label{composition_results_neighbour_fn}
\end{table*}
\subsubsection*{Composition by \emph{Union} or \emph{Intersection}}
\noindent Both model types in this study support composition by \emph{union} as well as composition by \emph{intersection}. In untyped models, composition by \emph{union} and composition by \emph{intersection} can be achieved by pointwise addition and pointwise multiplication respectively. The major difference between composition in the \APT~model and the untyped model is that in the former, composition is not commutative due to offsetting the modifier in a dependency relation (see section~\ref{background}). Blacoe and Lapata~\shortcite{Blacoe_2012} showed that an intersective composition function such as pointwise multiplication represents a competitive and robust approach in comparison to more sophisticated composition methods. For the final set of experiments on the Mitchell and Lapata~\shortcite{Mitchell_2010} dataset, we present results the \APT~model and the untyped model, using composition by \emph{union} and composition by \emph{intersection}, with and without distributional inference. We compare our models with the best performing untyped VSMs of Mitchell and Lapata~\shortcite{Mitchell_2010}, and Blacoe and Lapata~\shortcite{Blacoe_2012}, the best performing \APT~model of Weir et al.~\shortcite{Weir_2016}, as well as with the recently published state-of-the-art methods by Hashimoto et al.~\shortcite{Hashimoto_2014}, and Wieting et al.~\shortcite{Wieting_2015}, who are using neural network based approaches. For our models, we use the static top $n$ approach as neighbour retrieval function and tune the remaining parameters, the SPPMI shift $k$ ($1$, $5$, $10$, $40$, $100$) and the number of neighbours ($10$, $30$, $50$, $100$, $500$, $1000$, $5000$), for both model types, and the sliding window size for the untyped VSM ($1$, $2$, $5$), on the development portion of the Mitchell and Lapata~\shortcite{Mitchell_2010} dataset. We keep the vector configuration ($k$ and window size) fixed for all phrase types and only tune the number of neighbours used for DI individually. The best vector configuration for the \APT~model is achieved with $k=10$ and for the untyped VSM with $k=1$. For composition by \emph{intersection} best performance on the dev set was achieved with $1000$ neighbours for ANs, $10$ for NNs and $50$ for VOs with DI. For composition by \emph{union}, top performance was obtained with $100$ neighbours for ANs, $30$ neighbours for NNs and $50$ for VOs. The best results for the untyped model on the dev set are achieved with a symmetric window size of $1$ and using $5000$ neighbours for ANs, $10$ for NNs and $1000$ for VOs with composition by pointwise multiplication, and $30$ neighbours for ANs, $5000$ for NNs and $5000$ for VOs for composition by pointwise addition. The validated numbers of neighbours on the development set show that the problem of missing information appears to be more severe for semantic composition than for word similarity tasks. Even though a neighbour at rank $1000$ or lower does not appear to have a close relationship to the target word, it still can contribute useful co-occurrence information not observed in the original vector.\\
\begin{table*}[!htb]
\centering
\small
\resizebox{\textwidth}{!}{
\begin{tabular}{ l | c | c | c | c }
\bf{Model} & \bf{Adjective-Noun} & \bf{Noun-Noun} & \bf{Verb-Object} & \bf{Average}\\ \hline
\APT~-- \emph{union} & $0.45$ $(0.45)$ & $0.45$ $(0.43)$ & $0.38$ $(0.37)$ & $0.43$ $(0.42)$\\
\APT~-- \emph{intersect} & \underline{$0.50$} $(0.38)$ & \underline{$0.49$} $(0.44)$ & \underline{$0.43$} $(0.36)$ & \underline{$0.47$} $(0.39)$\\
Untyped VSM -- \emph{addition} & $0.46$ $(0.46)$ & $0.40$ $(0.41)$ & $0.38$ $(0.33)$ & $0.41$ $(0.40)$\\
Untyped VSM -- \emph{multiplication} & $0.46$ $(0.42)$ & $0.48$ $(0.45)$ & $0.40$ $(0.39)$ & $0.45$ $(0.42)$\\ \hline
Mitchell and Lapata~\shortcite{Mitchell_2010} (untyped VSM \& \emph{multiplication}) & $0.46$ & $0.49$ & $0.37$ & $0.44$\\
Blacoe and Lapata~\shortcite{Blacoe_2012} (untyped VSM \& \emph{multiplication}) & $0.48$ & $\bf{\underline{0.50}}$ & $0.35$ & $0.44$\\
Hashimoto et al.~\shortcite{Hashimoto_2014} (PAS-CLBLM \& $\text{\emph{Add}}_{nl}$) & $\bf{\underline{0.52}}$ & $0.46$ & $0.45$ & $\bf{\underline{0.48}}$\\
Wieting et al.~\shortcite{Wieting_2015} (Paragram word embeddings \& \emph{RNN}) & $0.51$ & $0.40$ & $\bf{\underline{0.50}}$ & $0.47$\\
Weir et al.~\shortcite{Weir_2016} (\APT~\& \emph{union}) & $0.45$ & $0.42$ & $0.42$ & $0.43$\\ \hline
\end{tabular}}
\captionsetup{font=tiny}
\caption{Results for the Mitchell and Lapata~\shortcite{Mitchell_2010} dataset. Results in brackets denote the performance of the respective models without the use of distributional inference. Underlined means best within group, boldfaced means best overall.}
\label{ml2010_results_overview}
\end{table*}
Table~\ref{ml2010_results_overview} shows that composition by \emph{intersection} with distributional inference considerably improves upon the best results for \APT~models without distributional inference and for untyped count-based models, and is competitive with the state-of-the-art neural network based models of Hashimoto et al.~\shortcite{Hashimoto_2014} and Wieting et al.~\shortcite{Wieting_2015}. Distributional inference also improves upon the performance of an untyped VSM where composition by pointwise multiplication is outperforming the models of Mitchell and Lapata~\shortcite{Mitchell_2010}, and Blacoe and Lapata~\shortcite{Blacoe_2012}. Table~\ref{ml2010_results_overview} furthermore shows that DI has a smaller effect on the \APT~model based on composition by \emph{union} and the untyped model based on composition by pointwise addition. The reason, as pointed out in the discussion for Table~\ref{composition_results_qualitative}, is that the composition function has no disambiguating effect and thus cannot eliminate unrelated neighbours introduced by distributional inference. An intersective composition function on the other hand is able to perform the disambiguation locally in any given phrasal context. This furthermore suggests that for the \APT~model it is not necessary to explicitly model different word senses in separate vectors, as composition by \emph{intersection} is able to disambiguate any word in context individually. Unlike the models of Hashimoto et al.~\shortcite{Hashimoto_2014} and Wieting et al.~\shortcite{Wieting_2015}, the elementary word representations, as well as the representations for composed phrases and the composition process in our models are fully interpretable\footnote{We release the \APT~vectors and our code at \url{https://github.com/tttthomasssss/apt-toolkit}.}. 

\section{Conclusion and Future Work}
One of the major challenges in count-based models is dealing with extreme sparsity and missing information. This paper contributes a number of findings relating to this challenge, in particular a simple unsupervised algorithm for enriching sparse word representations by leveraging its distributional neighbourhood. We have demonstrated its benefit to typed and untyped vector space models on a range of word similarity datasets. We have shown that distributional inference improves the performance of typed and untyped VSMs for semantic composition and that our \APT~model is competitive with the state-of-the-art for adjective-noun, noun-noun and verb-object compositions while being fully interpretable. With our method, we are able to bridge the gap in performance between low-dimensional non-interpretable and high-dimensional interpretable representations. Lastly, we have investigated the different behaviour of composition by \emph{union} and composition by \emph{intersection} and have shown that an \emph{intersective} composition function, together with distributional inference, has the ability to locally disambiguate the meaning of a phrase.\\\indent
In future work we aim to scale our approach to semantic composition with distributional inference to longer phrases and full sentences. We furthermore plan to investigate whether the number of neighbours required for improving elementary vector representations remains as high for other compositional tasks and longer phrases as in this study.
\label{conclusion}
\clearpage

\section*{Acknowledgments}
This work was funded by UK EPSRC project EP/IO37458/1 \emph{``A Unified Model of Compositional and Distributional Compositional Semantics: Theory and Applications"}. We would like to thank Miroslav Batchkarov for valuable discussions on earlier drafts of this paper and our anonymous reviewers for their helpful comments.

\bibliography{emnlp2016}

\begin{thebibliography}{}

\bibitem[\protect\citename{Agirre \bgroup et al.\egroup }2009]{Agirre_2009}
Eneko Agirre, Enrique Alfonseca, Keith Hall, Jana Kravalova, Marius Pasca, and
  Aitor Soroa.
\newblock 2009.
\newblock A study on similarity and relatedness using distributional and
  wordnet-based approaches.
\newblock In {\em Proceedings of NAACL-HLT}, pages 19--27, Boulder, Colorado,
  June. Association for Computational Linguistics.

\bibitem[\protect\citename{Baroni and Lenci}2010]{Baroni_2010b}
Marco Baroni and Alessandro Lenci.
\newblock 2010.
\newblock Distributional memory: A general framework for corpus-based
  semantics.
\newblock {\em Computational Linguistics}, 36(4):673--721, December.

\bibitem[\protect\citename{Baroni and Lenci}2011]{Baroni_2011}
Marco Baroni and Alessandro Lenci.
\newblock 2011.
\newblock How we blessed distributional semantic evaluation.
\newblock In {\em Proceedings of GEMS Workshop}, GEMS '11, pages 1--10,
  Stroudsburg, PA, USA. Association for Computational Linguistics.

\bibitem[\protect\citename{Baroni and Zamparelli}2010]{Baroni_2010a}
Marco Baroni and Roberto Zamparelli.
\newblock 2010.
\newblock Nouns are vectors, adjectives are matrices: Representing
  adjective-noun constructions in semantic space.
\newblock In {\em Proceedings of EMNLP}, pages 1183--1193, Cambridge, MA,
  October. Association for Computational Linguistics.

\bibitem[\protect\citename{Blacoe and Lapata}2012]{Blacoe_2012}
William Blacoe and Mirella Lapata.
\newblock 2012.
\newblock A comparison of vector-based representations for semantic
  composition.
\newblock In {\em Proceedings of EMNLP}, pages 546--556, Jeju Island, Korea,
  July. Association for Computational Linguistics.

\bibitem[\protect\citename{Bruni \bgroup et al.\egroup }2014]{Bruni_2014}
Elia Bruni, Nam~Khanh Tran, and Marco Baroni.
\newblock 2014.
\newblock Multimodal distributional semantics.
\newblock {\em J. Artif. Int. Res.}, 49(1):1--47, January.

\bibitem[\protect\citename{Bullinaria and Levy}2007]{Bullinaria_2007}
John~A. Bullinaria and Joseph~P. Levy.
\newblock 2007.
\newblock Extracting semantic representations from word co-occurrence
  statistics: A computational study.
\newblock {\em Behavior Research Methods}, pages 510--526.

\bibitem[\protect\citename{Bullinaria and Levy}2012]{Bullinaria_2012}
John~A. Bullinaria and Joseph~P. Levy.
\newblock 2012.
\newblock Extracting semantic representations from word co-occurrence
  statistics: stop-lists, stemming, and svd.
\newblock {\em Behavior Research Methods}, 44(3):890--907.

\bibitem[\protect\citename{Church and Hanks}1990]{Church_1990}
Kenneth~Ward Church and Patrick Hanks.
\newblock 1990.
\newblock Word association norms, mutual information, and lexicography.
\newblock {\em Computational Linguistics}, 16(1):22--29, March.

\bibitem[\protect\citename{Coecke \bgroup et al.\egroup }2010]{Coecke_2010}
Bob Coecke, Mehrnoosh Sadrzadeh, and Stephen Clark.
\newblock 2010.
\newblock Mathematical foundations for a compositional distributional model of
  meaning.
\newblock {\em CoRR}, abs/1003.4394.

\bibitem[\protect\citename{Curran}2004]{Curran_2004}
James Curran.
\newblock 2004.
\newblock {\em From Distributional to Semantic Similarity}.
\newblock {Ph.D.} thesis, University of Edinburgh.

\bibitem[\protect\citename{Dagan \bgroup et al.\egroup }1994]{Dagan_1994}
Ido Dagan, Fernando Pereira, and Lillian Lee.
\newblock 1994.
\newblock Similarity-based estimation of word cooccurrence probabilities.
\newblock In {\em Proceedings of ACL}, pages 272--278, Las Cruces, New Mexico,
  USA, June. Association for Computational Linguistics.

\bibitem[\protect\citename{Dagan \bgroup et al.\egroup }1997]{Dagan_1997}
Ido Dagan, Lillian Lee, and Fernando Pereira.
\newblock 1997.
\newblock Similarity-based methods for word sense disambiguation.
\newblock In {\em Proceedings of ACL}, pages 56--63, Madrid, Spain, July.
  Association for Computational Linguistics.

\bibitem[\protect\citename{de Marneffe \bgroup et al.\egroup
  }2014]{deMarneffe_2014}
Marie-Catherine de~Marneffe, Timothy Dozat, Natalia Silveira, Katri Haverinen,
  Filip Ginter, Joakim Nivre, and Christopher~D. Manning.
\newblock 2014.
\newblock Universal stanford dependencies: A cross-linguistic typology.
\newblock In {\em Proceedings of LREC}, pages 4585--4592, Reykjavik, Iceland,
  May. European Language Resources Association (ELRA).
\newblock ACL Anthology Identifier: L14-1045.

\bibitem[\protect\citename{Deerwester \bgroup et al.\egroup
  }1990]{Deerwester_1990}
Scott Deerwester, Susan~T. Dumais, George~W. Furnas, Thomas~K. Landauer, and
  Richard Harshman.
\newblock 1990.
\newblock Indexing by latent semantic analysis.
\newblock {\em J. Amer. Soc. Inf. Sci.}, 41(6):391--407.

\bibitem[\protect\citename{Erk and Pad\'{o}}2008]{Erk_2008}
Katrin Erk and Sebastian Pad\'{o}.
\newblock 2008.
\newblock A structured vector space model for word meaning in context.
\newblock In {\em Proceedings of EMNLP}, pages 897--906, Honolulu, Hawaii,
  October. Association for Computational Linguistics.

\bibitem[\protect\citename{Fellbaum}1998]{Fellbaum_1998}
Christiane Fellbaum, editor.
\newblock 1998.
\newblock {\em {WordNet: an electronic lexical database}}.
\newblock MIT Press.

\bibitem[\protect\citename{Finkelstein \bgroup et al.\egroup
  }2001]{Finkelstein_2001}
Lev Finkelstein, Evgeniy Gabrilovich, Yossi Matias, Ehud Rivlin, Zach Solan,
  Gadi Wolfman, and Eytan Ruppin.
\newblock 2001.
\newblock Placing search in context: The concept revisited.
\newblock In {\em Proceedings of WWW}, WWW '01, pages 406--414, New York, NY,
  USA. ACM.

\bibitem[\protect\citename{Frege}1884]{Frege_1884}
Gottlob Frege.
\newblock 1884.
\newblock {\em Die Grundlagen der Arithmetik: Eine logisch mathematische
  Untersuchung {\"u}ber den Begriff der Zahl}.
\newblock W. Koebner.

\bibitem[\protect\citename{Grefenstette \bgroup et al.\egroup
  }2013]{Grefenstette_2013}
Edward Grefenstette, Georgiana Dinu, Yao-Zhong Zhang, Mehrnoosh Sadrzadeh, and
  Marco Baroni.
\newblock 2013.
\newblock Multi-step regression learning for compositional distributional
  semantics.
\newblock {\em Proceedings of IWCS}.

\bibitem[\protect\citename{Grefenstette}1994]{Grefenstette_1994}
Gregory Grefenstette.
\newblock 1994.
\newblock {\em Explorations in Automatic Thesaurus Discovery}.
\newblock Kluwer Academic Publishers, Norwell, MA, USA.

\bibitem[\protect\citename{Hashimoto \bgroup et al.\egroup
  }2014]{Hashimoto_2014}
Kazuma Hashimoto, Pontus Stenetorp, Makoto Miwa, and Yoshimasa Tsuruoka.
\newblock 2014.
\newblock Jointly learning word representations and composition functions using
  predicate-argument structures.
\newblock In {\em Proceedings of EMNLP}, pages 1544--1555, Doha, Qatar,
  October. Association for Computational Linguistics.

\bibitem[\protect\citename{Hill \bgroup et al.\egroup }2015]{Hill_2015}
Felix Hill, Roi Reichart, and Anna Korhonen.
\newblock 2015.
\newblock Simlex-999: Evaluating semantic models with (genuine) similarity
  estimation.
\newblock {\em Computational Linguistics}, 41(4):665--695, December.

\bibitem[\protect\citename{Lapesa and Evert}2014]{Lapesa_2014}
Gabriella Lapesa and Stefan Evert.
\newblock 2014.
\newblock A large scale evaluation of distributional semantic models:
  Parameters, interactions and model selection.
\newblock {\em TACL}, 2:531--545.

\bibitem[\protect\citename{Le and Zuidema}2015]{Le_2015}
Phong Le and Willem Zuidema.
\newblock 2015.
\newblock The forest convolutional network: Compositional distributional
  semantics with a neural chart and without binarization.
\newblock In {\em Proceedings of EMNLP}, pages 1155--1164, Lisbon, Portugal,
  September. Association for Computational Linguistics.

\bibitem[\protect\citename{Levy and Goldberg}2014]{Levy_2014c}
Omer Levy and Yoav Goldberg.
\newblock 2014.
\newblock Neural word embedding as implicit matrix factorization.
\newblock In {\em Proceedings of NIPS}, pages 2177--2185.

\bibitem[\protect\citename{Levy \bgroup et al.\egroup }2015]{Levy_2015}
Omer Levy, Yoav Goldberg, and Ido Dagan.
\newblock 2015.
\newblock Improving distributional similarity with lessons learned from word
  embeddings.
\newblock {\em TACL}, 3:211--225.

\bibitem[\protect\citename{Lin}1998]{Lin_1998}
Dekang Lin.
\newblock 1998.
\newblock Automatic retrieval and clustering of similar words.
\newblock In {\em Proceedings of ACL}, pages 768--774, Montreal, Quebec,
  Canada, August. Association for Computational Linguistics.

\bibitem[\protect\citename{Lund and Burgess}1996]{Lund_1996}
Kevin Lund and Curt Burgess.
\newblock 1996.
\newblock Producing high-dimensional semantic spaces from lexical
  co-occurrence.
\newblock {\em Behavior Research Methods, Instruments, \& Computers},
  28(2):203--208.

\bibitem[\protect\citename{Manning \bgroup et al.\egroup }2014]{Manning_2014}
Christopher~D. Manning, Mihai Surdeanu, John Bauer, Jenny Finkel, Steven~J.
  Bethard, and David McClosky.
\newblock 2014.
\newblock The {Stanford} {CoreNLP} natural language processing toolkit.
\newblock In {\em Proceedings of ACL - System Demonstrations}, pages 55--60.

\bibitem[\protect\citename{Mikolov \bgroup et al.\egroup }2013]{Mikolov_2013a}
Tomas Mikolov, Ilya Sutskever, Kai Chen, Greg~S Corrado, and Jeff Dean.
\newblock 2013.
\newblock Distributed representations of words and phrases and their
  compositionality.
\newblock In {\em Proceedings of NIPS}, pages 3111--3119.

\bibitem[\protect\citename{Mitchell and Lapata}2008]{Mitchell_2008}
Jeff Mitchell and Mirella Lapata.
\newblock 2008.
\newblock Vector-based models of semantic composition.
\newblock In {\em In Proceedings of ACL-08: HLT}, pages 236--244.

\bibitem[\protect\citename{Mitchell and Lapata}2010]{Mitchell_2010}
Jeff Mitchell and Mirella Lapata.
\newblock 2010.
\newblock Composition in distributional models of semantics.
\newblock {\em Cognitive Science}, 34(8):1388--1429.

\bibitem[\protect\citename{Mou \bgroup et al.\egroup }2015]{Mou_2015}
Lili Mou, Hao Peng, Ge~Li, Yan Xu, Lu~Zhang, and Zhi Jin.
\newblock 2015.
\newblock Discriminative neural sentence modeling by tree-based convolution.
\newblock In {\em Proceedings of EMNLP}, pages 2315--2325, Lisbon, Portugal,
  September. Association for Computational Linguistics.

\bibitem[\protect\citename{Niwa and Nitta}1994]{Niwa_1994}
Yoshiki Niwa and Yoshihiko Nitta.
\newblock 1994.
\newblock Co-occurrence vectors from corpora vs. distance vectors from
  dictionaries.
\newblock In {\em Proceedings of Coling}, COLING '94, pages 304--309,
  Stroudsburg, PA, USA. Association for Computational Linguistics.

\bibitem[\protect\citename{Pad\'{o} and Lapata}2007]{Pado_2007}
Sebastian Pad\'{o} and Mirella Lapata.
\newblock 2007.
\newblock Dependency-based construction of semantic space models.
\newblock {\em Computational Linguistics}, 33(2):161--199.

\bibitem[\protect\citename{Pad\'{o} \bgroup et al.\egroup }2013]{Pado_2013}
Sebastian Pad\'{o}, Jan \v{S}najder, and Britta Zeller.
\newblock 2013.
\newblock Derivational smoothing for syntactic distributional semantics.
\newblock In {\em Proceedings of ACL}, pages 731--735, Sofia, Bulgaria, August.
  Association for Computational Linguistics.

\bibitem[\protect\citename{Paperno \bgroup et al.\egroup }2014]{Paperno_2014}
Denis Paperno, Nghia~The Pham, and Marco Baroni.
\newblock 2014.
\newblock A practical and linguistically-motivated approach to compositional
  distributional semantics.
\newblock In {\em Proceedings of ACL}, pages 90--99, Baltimore, Maryland, June.
  Association for Computational Linguistics.

\bibitem[\protect\citename{Parzen}1962]{Parzen_1962}
Emanuel Parzen.
\newblock 1962.
\newblock On estimation of a probability density function and mode.
\newblock {\em Ann. Math. Statist.}, 33(3):1065--1076, 09.

\bibitem[\protect\citename{Pennington \bgroup et al.\egroup
  }2014]{Pennington_2014}
Jeffrey Pennington, Richard Socher, and Christopher Manning.
\newblock 2014.
\newblock Glove: Global vectors for word representation.
\newblock In {\em Proceedings of EMNLP}, pages 1532--1543, Doha, Qatar,
  October. Association for Computational Linguistics.

\bibitem[\protect\citename{Sahlgren}2006]{Sahlgren_2006}
Magnus Sahlgren.
\newblock 2006.
\newblock {\em {The Word-space model}}.
\newblock {Ph.D.} thesis, University of Stockholm (Sweden).

\bibitem[\protect\citename{Scheible \bgroup et al.\egroup }2013]{Scheible_2013}
Silke Scheible, Sabine Schulte~im Walde, and Sylvia Springorum.
\newblock 2013.
\newblock Uncovering distributional differences between synonyms and antonyms
  in a word space model.
\newblock In {\em Proceedings of IJCNLP}, pages 489--497, Nagoya, Japan,
  October. Asian Federation of Natural Language Processing.

\bibitem[\protect\citename{Socher \bgroup et al.\egroup }2012]{Socher_2012}
Richard Socher, Brody Huval, Christopher~D. Manning, and Andrew~Y. Ng.
\newblock 2012.
\newblock Semantic compositionality through recursive matrix-vector spaces.
\newblock In {\em Proceedings of EMNLP}, pages 1201--1211, Jeju Island, Korea,
  July. Association for Computational Linguistics.

\bibitem[\protect\citename{Tai \bgroup et al.\egroup }2015]{Tai_2015}
Kai~Sheng Tai, Richard Socher, and Christopher~D. Manning.
\newblock 2015.
\newblock Improved semantic representations from tree-structured long
  short-term memory networks.
\newblock In {\em Proceedings of ACL}, pages 1556--1566, Beijing, China, July.
  Association for Computational Linguistics.

\bibitem[\protect\citename{Thater \bgroup et al.\egroup }2010]{Thater_2010}
Stefan Thater, Hagen F\"{u}rstenau, and Manfred Pinkal.
\newblock 2010.
\newblock Contextualizing semantic representations using syntactically enriched
  vector models.
\newblock In {\em Proceedings of ACL}, pages 948--957, Uppsala, Sweden, July.
  Association for Computational Linguistics.

\bibitem[\protect\citename{Thater \bgroup et al.\egroup }2011]{Thater_2011}
Stefan Thater, Hagen F\"{u}rstenau, and Manfred Pinkal.
\newblock 2011.
\newblock Word meaning in context: A simple and effective vector model.
\newblock In {\em Proceedings of IJCNLP}, pages 1134--1143, Chiang Mai,
  Thailand, November. Asian Federation of Natural Language Processing.

\bibitem[\protect\citename{Turney and Pantel}2010]{Turney_2010}
Peter~D. Turney and Patrick Pantel.
\newblock 2010.
\newblock From frequency to meaning: Vector space models of semantics.
\newblock {\em J. Artif. Int. Res.}, 37(1):141--188, January.

\bibitem[\protect\citename{Weeds \bgroup et al.\egroup }2014]{Weeds_2014}
Julie Weeds, David Weir, and Jeremy Reffin.
\newblock 2014.
\newblock Distributional composition using higher-order dependency vectors.
\newblock In {\em Proceedings of the 2nd Workshop on Continuous Vector Space
  Models and their Compositionality}, pages 11--20, Gothenburg, Sweden, April.
  Association for Computational Linguistics.

\bibitem[\protect\citename{Weir \bgroup et al.\egroup }2016]{Weir_2016}
David Weir, Julie Weeds, Jeremy Reffin, and Thomas Kober.
\newblock 2016.
\newblock Aligning packed dependency trees: a theory of composition for
  distributional semantics.
\newblock {\em Computational Linguistics}, in press.

\bibitem[\protect\citename{Wieting \bgroup et al.\egroup }2015]{Wieting_2015}
John Wieting, Mohit Bansal, Kevin Gimpel, and Karen Livescu.
\newblock 2015.
\newblock From paraphrase database to compositional paraphrase model and back.
\newblock {\em TACL}, 3:345--358.

\bibitem[\protect\citename{Wilson}2015]{Wilson_2015}
Benjamin Wilson.
\newblock 2015.
\newblock The unknown perils of mining wikipedia.
\newblock
  https://blog.lateral.io/2015/06/the-unknown-perils-of-mining-wikipedia/,
  June.

\bibitem[\protect\citename{Zanzotto \bgroup et al.\egroup }2015]{Zanzotto_2015}
Fabio~Massimo Zanzotto, Lorenzo Ferrone, and Marco Baroni.
\newblock 2015.
\newblock Squibs: When the whole is not greater than the combination of its
  parts: A "decompositional" look at compositional distributional semantics.
\newblock {\em Computational Linguistics}, 41(1):165--173.

\end{thebibliography}
\bibliographystyle{emnlp2016}

\end{document}